\documentclass{article}

\PassOptionsToPackage{numbers, compress}{natbib}
\usepackage[dblblindworkshop,final]{neurips_2025}

\workshoptitle{MATH-AI}

\usepackage[utf8]{inputenc} 
\usepackage[T1]{fontenc}    
\usepackage{hyperref}       
\usepackage{url}            
\usepackage{booktabs}       
\usepackage{amsfonts}       
\usepackage{nicefrac}       
\usepackage{microtype}      
\usepackage{xcolor}         
\usepackage{array} 
\usepackage{caption}
\usepackage[most]{tcolorbox}
\usepackage{enumitem} 
\usepackage{graphicx}
\usepackage{wrapfig}
\usepackage{mathtools}
\usepackage{amsmath,amssymb}
\usepackage[ruled,vlined]{algorithm2e} 
\usepackage{subcaption}
\SetAlgoNlRelativeSize{0} 
\SetNlSty{}{}{:}          
\SetAlgoNlRelativeSize{-1}
\SetAlgoNlRelativeSize{0}
\SetAlgoNlRelativeSize{-2}
\DontPrintSemicolon
\SetAlgoSkip{medskip}

\usepackage{algorithm2e} 

\title{Limits of Generalization in RLVR: Two Case Studies in Mathematical Reasoning}

\author{%
  Md Tanvirul Alam \\
  Rochester Institute of Technology\\
  Rochester, NY, USA \\
  \texttt{ma8235@rit.edu} \\
  \And
  Nidhi Rastogi \\
  Rochester Institute of Technology\\
  Rochester, NY, USA \\
  \texttt{nxrvse@rit.edu} \\
}

\begin{document}

\maketitle

\begin{abstract}


Mathematical reasoning is a central challenge for large language models (LLMs), requiring not only correct answers but also faithful reasoning processes. Reinforcement Learning with Verifiable Rewards (RLVR) has emerged as a promising approach for enhancing such capabilities; however, its ability to foster genuine reasoning remains unclear. We investigate RLVR on two combinatorial problems with fully verifiable solutions: \emph{Activity Scheduling} and the \emph{Longest Increasing Subsequence}, using carefully curated datasets with unique optima. Across multiple reward designs, we find that RLVR improves evaluation metrics but often by reinforcing superficial heuristics rather than acquiring new reasoning strategies. These findings highlight the limits of RLVR generalization, emphasizing the importance of benchmarks that disentangle genuine mathematical reasoning from shortcut exploitation and provide faithful measures of progress. Code available at \url{https://github.com/xashru/rlvr-seq-generalization}.

\end{abstract}

\section{Introduction}

Large language models (LLMs) have recently advanced rapidly on mathematical and programming benchmarks~\cite{guo2025deepseek,jaech2024openai,wu2024reasoning,comanici2025gemini,yang2025qwen3}. A key driver is \emph{Reinforcement Learning with Verifiable Rewards} (RLVR), which fine-tunes pretrained models against automatically checkable signals such as exact answers or unit tests~\cite{guo2025deepseek,yu2025dapo}. This paradigm eliminates reliance on human annotation, enabling scalable training on large problem sets and delivering consistent gains on challenging reasoning tasks~\cite{el2025competitive}.

Despite strong empirical gains, the nature of RLVR’s improvements remains unclear. Studies show it often boosts accuracy while reducing exploration~\cite{wu2025invisible,cui2025entropy,yue2025does,wang2025beyond}, with base model ability acting as a ceiling~\cite{yue2025does,wu2025invisible}. Gains largely reflect re-weighting existing solutions, and reasoning coverage can even contract at larger sample sizes. Standard $Pass@K$ further over-credits “lucky” completions~\cite{wen2025reinforcement}, whereas stricter metrics like $CoT$-$Pass@K$ are more reliable but harder to scale. Reward design can also yield surprising effects: a single example can rival large-scale training~\cite{wang2025reinforcement}, and even spurious rewards can drive improvements in models with strong procedural biases~\cite{shao2025spurious}. Overall, RLVR appears to stabilize existing competencies rather than induce new reasoning strategies.

While prior studies have offered valuable insights, many rely on benchmarks where the correctness of reasoning is difficult to verify, making it unclear whether improvements reflect genuine mathematical competence or superficial pattern matching. We address this gap by focusing on two combinatorial problems with fully verifiable solutions: \emph{Activity Scheduling}, which admits a unique greedy optimum, and the \emph{Longest Increasing Subsequence} (LIS), which can be solved using dynamic programming. By constructing datasets where each instance has a single optimal sequence, we can precisely measure not only answer accuracy but also sequence fidelity and structural behaviors, such as correct sorting in Activity Scheduling. This verifiable setup provides a rigorous lens on RLVR, revealing when observed gains arise from heuristic shortcuts versus genuine reasoning strategies, and highlighting broader implications for the design of mathematical reasoning benchmarks.

\section{Experimental Setup}




\setlist[itemize]{leftmargin=1.2em, topsep=0.2em, itemsep=0.1em} 
\setlist[enumerate]{leftmargin=1.6em, topsep=0.2em, itemsep=0.1em} 
\begin{figure}[t] 
\centering 
\setlength{\tabcolsep}{6pt} 
\renewcommand{\arraystretch}{1.0} 
\begin{tabular}{@{}p{0.49\linewidth} p{0.49\linewidth}@{}}
\begin{tcolorbox}[colback=white,colframe=black!60,boxrule=0.5pt, sharp corners,enhanced jigsaw, title=Activity Scheduling  Example] 
\scriptsize 
Determine the largest subset of activities that can be scheduled without any overlaps. \\
\vspace{0.4em} 
\begin{tabular}{@{}rcc@{}} 
ID & Start & End\\ \hline 
1 & 06:09 & 07:24 \\ 
2 & 07:13 & 08:23 \\ 
3 & 07:29 & 09:28 \\ 
4 & 08:24 & 10:18 \\ 
5 & 04:48 & 06:14 \\ 
\end{tabular} 
\vspace{0.3em}  \\
\textbf{Ground truth:} \texttt{\string\ids\{5,2,4\}}, \texttt{\string\answer\{3\}} 
\end{tcolorbox} & 
\begin{tcolorbox}[colback=white,colframe=black!60,boxrule=0.5pt, sharp corners,enhanced jigsaw, title=LIS Example] 
\scriptsize 
Determine the longest strictly increasing subsequence (by VALUE) from the rows below (use the row IDs). \\
\vspace{0.4em} 
\begin{tabular}{@{}rc@{}} 
ID & Value\\ \hline 
1 & 797 \\ 
2 & 476 \\ 
3 & 335 \\ 
4 & 452 \\ 
5 & 606 \\ 
\end{tabular} 

\vspace{0.3em}

\textbf{Ground truth:} \texttt{\string\ids\{3,4,5\}}, \texttt{\string\answer\{3\}} 
\end{tcolorbox} \\ 
\end{tabular} 
\caption{Example question and ground-truth for Activity Scheduling (left) and LIS (right).} 
\label{fig:task-example} 
\end{figure}

\subsection{Tasks}
\textbf{Activity Scheduling.} 
Each activity $i$ has start and finish times $(s_i,f_i)$ with $s_i<f_i$, and intervals are half-open $[s_i,f_i)$. 
The goal is to select a maximum subset of non-overlapping activities~\cite{wikipedia_activity_selection}. 
Instances are constructed so that the greedy earliest-finish-time algorithm with deterministic tie-breaking yields the unique optimum, reported as IDs sorted by $f_i$ (ties by smaller $i$).  

\textbf{Longest Increasing Subsequence (LIS).} 
Given $a_1,\dots,a_n \in \mathbb{Z}$, find indices $1 \le i_1 < \cdots < i_k \le n$ maximizing $k$ with $a_{i_1}<\cdots<a_{i_k}$. 
Uniqueness is enforced via an $O(n^2)$ dynamic-programming count, and the LIS is reconstructed with patience sorting in $O(n\log n)$~\cite{wikipedia_lis}.  

\noindent Our generator enforces uniqueness of the optimal solution for both tasks, yielding a single ground-truth ID sequence with a fixed canonical reporting order (see Appendix~\ref{app:data-generator}).

\textbf{Dataset.} 
We generate $2000$ instances per task, half with hints, with sequence lengths $5$–$16$. 
To avoid leakage, train and test use disjoint length ranges, leaving $462$ test cases for Activity and $428$ for LIS; the remainder form the training set. 
Example questions and their corresponding ground truths are illustrated in Fig.~\ref{fig:task-example}, with detailed prompts provided in Fig.~\ref{fig:prompt} in Appendix.

\subsection{Reward Functions}
\label{sec:rewards}

For each instance $(x,y^\star)$ with ground-truth answer $a^\star$ and unique optimal sequence $s^\star=(i_1,\dots,i_L)$, the output $y$ is parsed into $\hat a(y)$ (from \texttt{\string\answer\{\dots\}}) and $\hat s(y)$ (from \texttt{\string\ids\{\dots\}}); if parsing fails, $\hat s(y)=\varnothing$. All rewards lie in $[0,1]$:

\medskip\noindent\textbf{(1) Answer-only.}  
This reward evaluates only the correctness of the final numeric answer:
\[
r_{\text{ans}}(y) \;=\; \mathbb{I}\!\left[\hat{a}(y) = a^\star\right],
\]
where $\mathbb{I}[\cdot]$ denotes the indicator function, equal to $1$ if the condition holds and $0$ otherwise.

\noindent\textbf{(2) Answer + Format (LIS only).}  
To stabilize behavior when the policy begins omitting reasoning, we introduce a small formatting bonus, applied only to the \textit{LIS} task(see \S\ref{lis-rl})).  
Define the format indicator
\[
\mathrm{fmt}(y) \;=\; \mathbb{I}\!\Bigl[
  \text{$y$ contains \texttt{<think>...</think>} and valid \texttt{\string\answer\{\dots\}}, \texttt{\string\ids\{\dots\}}}
\Bigr],
\]

and combine it with answer correctness using a mixing weight $\lambda = 0.1$:
\[
r_{\text{ans+fmt}}(y) \;=\; (1-\lambda)\, r_{\text{ans}}(y) \;+\; \lambda\, \mathrm{fmt}(y).
\]


\medskip\noindent\textbf{(3) Exact-IDs.}  
Rewards $1$ if and only if the predicted sequence matches the optimum:
\[
r_{\text{ids,exa}}(y) \;=\; \mathbb{I}\!\bigl[\hat{s}(y) = s^\star\bigr].
\]

\noindent\textbf{(4) Prefix-IDs.}  
This reward grants partial credit proportional to the length of the longest common prefix with the ground-truth sequence, while applying a small penalty if the predicted length is incorrect (clipped at $0$).  
Define
\[
m(y) \;=\; \max \bigl\{\, m \in \{0,\ldots,L\} : (\hat s_1,\ldots,\hat s_m) = (i_1,\ldots,i_m) \,\bigr\},
\]
and fix a length-penalty $\gamma = 0.1$.  
The reward is then
\[
r_{\text{ids,pre}}(y) \;=\; 
\max \Bigl\{\, 0,\ \frac{m(y)}{L} \;-\; 
\gamma\, \mathbb{I}\!\bigl[\,\hat{s}(y) = \varnothing \ \lor\ |\hat{s}(y)| \neq L \,\bigr] \Bigr\}.
\]

\noindent\textbf{(5) Sorting-Match (Activity only).}  
In the \textit{Activity Scheduling} task, sorting by finish time is the first step of the greedy algorithm. 
Interestingly, we observe that even without explicit instructions, model responses often begin with a sorted version of the input sequence, which can be extracted reliably (see \S\ref{sec:sorting_performance}). 
This motivates an auxiliary reward that checks whether the extracted sorted sequence matches the ground-truth sorted order of activities:

\[
r_{\text{sort}}(y) \;=\; \mathbb{I}\!\bigl[\, \hat{s}_{\text{sort}}(y) \,=\, s_{\text{sort}}^\star \,\bigr].
\]

where $\hat s_{\text{sort}}(y)$ is the sequence of activity IDs sorted as extracted from the model output, and $s_{\text{sort}}^\star$ is the canonical sorted sequence by increasing finish time (breaking ties by smaller ID).

\subsection{Training \& Evaluation}
We fine-tune \textit{Qwen2.5-7B-Instruct}~\cite{qwen2.5} with GRPO~\cite{shao2024deepseekmath,guo2025deepseek}  using the \texttt{verl} framework~\cite{sheng2024hybridflow}. 
Unless noted, max generation length is $T_{\max}=2048$ (extended to $7680$ for LIS with $r_{\text{ans+fmt}}$). 
Each PPO update uses $256$ prompts with $8$ rollouts from \texttt{vLLM}~\cite{kwon2023efficient}, trained for $20$ epochs ($120$ updates) at learning rate $10^{-6}$ and no KL penalty ($\beta_{\text{KL}}=0$). 
Training prompts match evaluation format.

\paragraph{Evaluation protocol.}
For each instance $x$ we draw $k=256$ samples 
$\{y_j\}_{j=1}^{256}$ with temperature $0.6$ and top-$p=0.95$, and parse each $y_j$ into $(\hat a(y_j), \hat s(y_j))$ as in the reward definitions. 
We evaluate two complementary notions of accuracy:
$\mathrm{Acc}_{\text{ans}}$ (correctness of the reported cardinality/length) and
$\mathrm{Acc}_{\text{ids}}$ (exact match of the ID sequence).
Both are measured under \emph{Pass@k} and \emph{Self-consistency (SC)}:

\begin{itemize}
  \item \textbf{Pass@k}~\cite{yue2025does}.  
  Success under a $k$-sample budget, defined by whether at least one of the $k$ generations is correct:

\[
\mathrm{Pass@k}_{\mathrm{ans}}(x) = \mathbb{I}\!\bigl[\exists\, j \le k:\ \hat{a}(y_j) = a^\star \bigr], 
\qquad
\mathrm{Pass@k}_{\mathrm{ids}}(x) = \mathbb{I}\!\bigl[\exists\, j \le k:\ \hat{s}(y_j) = s^\star \bigr].
\]

  \item \textbf{Self-consistency (SC)}~\cite{wang2022self}.  
  Agreement between the majority prediction aggregated over $k$ generations and the ground truth:
  \[
  \tilde a_k(x) := \operatorname{mode}\{\hat a(y_j)\}_{j=1}^k, 
  \qquad
  \tilde s_k(x) := \operatorname{mode}\{\hat s(y_j)\}_{j=1}^k,
  \]
  with deterministic tie-breaking (numerically smallest for answers; lexicographic for sequences). We report
\[
\mathrm{SC}_{\mathrm{ans}}(x;k) = \mathbb{I}\!\bigl[\tilde{a}_k(x) = a^\star \bigr],
\qquad
\mathrm{SC}_{\mathrm{ids}}(x;k) = \mathbb{I}\!\bigl[\tilde{s}_k(x) = s^\star \bigr].
\]

\end{itemize}

All metrics are averaged over the test set. 
Because each instance admits a unique ground-truth answer $a^\star$ and sequence $s^\star$, these definitions are unambiguous. 
We plot the full curves $\{\mathrm{Pass@k}\}_{k=1}^{256}$ and $\{\mathrm{SC}(\cdot;k)\}_{k=1}^{256}$, and also report their values at $k=256$.

\section{Results \& Analysis}

\subsection{Training with Exact Answer Reward}
\begin{figure}[t]
    \centering
    \includegraphics[width=\linewidth]{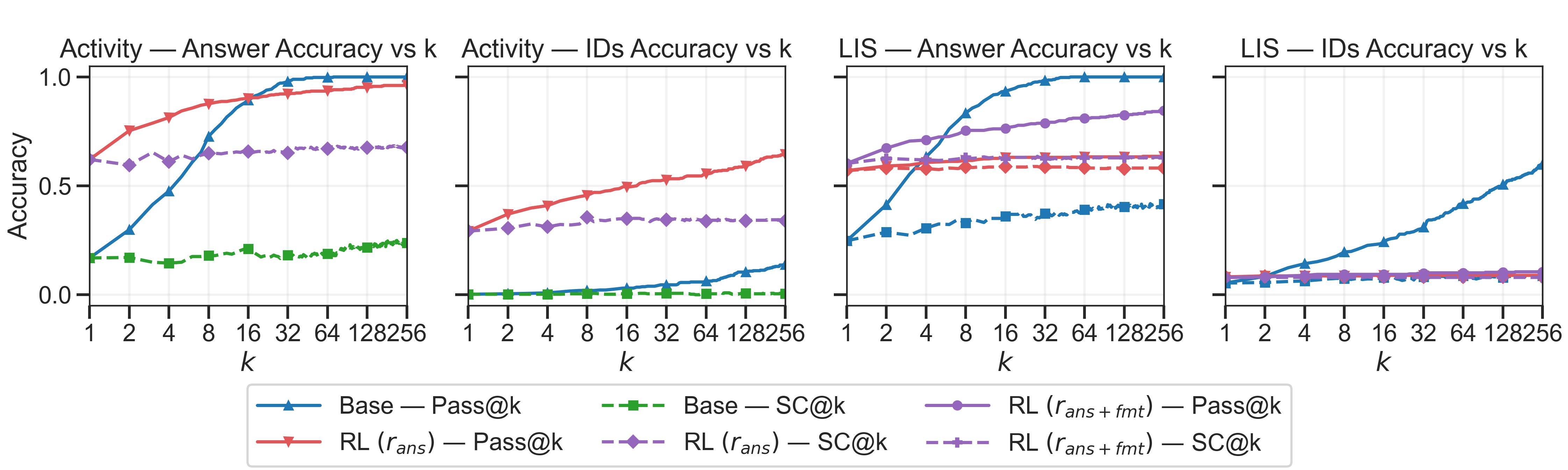}

\caption{Performance comparison of Base, RL($r_{\text{ans}}$), and RL($r_{\text{ans+fmt}}$) models on the \textit{Activity} and \textit{LIS} task with the Qwen2.5-7B model.}

    \label{fig:actlis_ans}

\end{figure}

Fig.~\ref{fig:actlis_ans} compares the base model and the RLVR-trained policy on \textit{Activity Scheduling} and LIS tasks.

\textbf{Activity Scheduling.} Since the target is a small integer (maximum schedule length $16$), the base model quickly saturates to \textit{Pass@k} $\approx 1.0$ as $k$ increases. While the RLVR-trained policy attains slightly lower \textit{Pass@k}$\,$ at large $k$, it achieves much higher self-consistency (about $0.68$ vs.\ $\sim\!0.24$ at $k{=}256$), indicating more stable predictions across samples. Under exact sequence ID matching, RLVR substantially outperforms the base model: at $k{=}256$, RLVR reaches $\mathrm{Pass@k} \approx 0.64$ compared to $0.14$, with $\mathrm{SC} \approx 0.34$ compared to $0.004$, reflecting a clear improvement in sequence-level fidelity. These results support prior findings that answer-only \textit{Pass@k} can overstate model capability~\cite{wen2025reinforcement}, whereas $\mathrm{Acc}_{\text{ids}}$ and SC provide more faithful measures of reasoning quality. RLVR improves both by reinforcing verified reasoning trajectories~\cite{wen2025reinforcement}.

\textbf{LIS.}
\label{lis-rl}
On \textit{LIS}, training with the answer-only reward $r_{\text{ans}}$ rapidly collapses intermediate reasoning: after just a few PPO updates, the policy drops its chain of thought and outputs terse final answers, reflected in a sharp decline in mean response length (Appendix~\ref{app:lismore}). Adding a format component ($r_{\text{ans+fmt}}$) mitigates this, keeping response lengths closer to the base model (Fig.~\ref{fig:response_len}). As shown in Figure~\ref{fig:actlis_ans}, both RLVR-trained models achieve higher SC on answer than the base model; for $r_{\text{ans}}$, the \textit{Pass@k} and SC curves nearly overlap, whereas $r_{\text{ans+fmt}}$ underperforms the base model at larger $k$. However, under exact sequence ID evaluation, both RLVR-trained policies remain weak, with low \textit{Pass@k} and SC compared to the base model, in contrast to the \textit{Activity} task, where RLVR improved both metrics.

\begin{tcolorbox}[colback=gray!10, colframe=black, title=Takeaway 1]
RLVR improves answer-level generalization on both \textit{Activity Scheduling} and \textit{LIS}. 
However, only in \textit{Activity Scheduling} do we observe reasoning gains, while on \textit{LIS}, improvements stem from superficial heuristics or formatting strategies. 
Thus, RLVR can yield apparent task generalization without strengthening the underlying reasoning process.
\end{tcolorbox}







\begin{figure}[t]
    \centering
    \includegraphics[width=\linewidth]{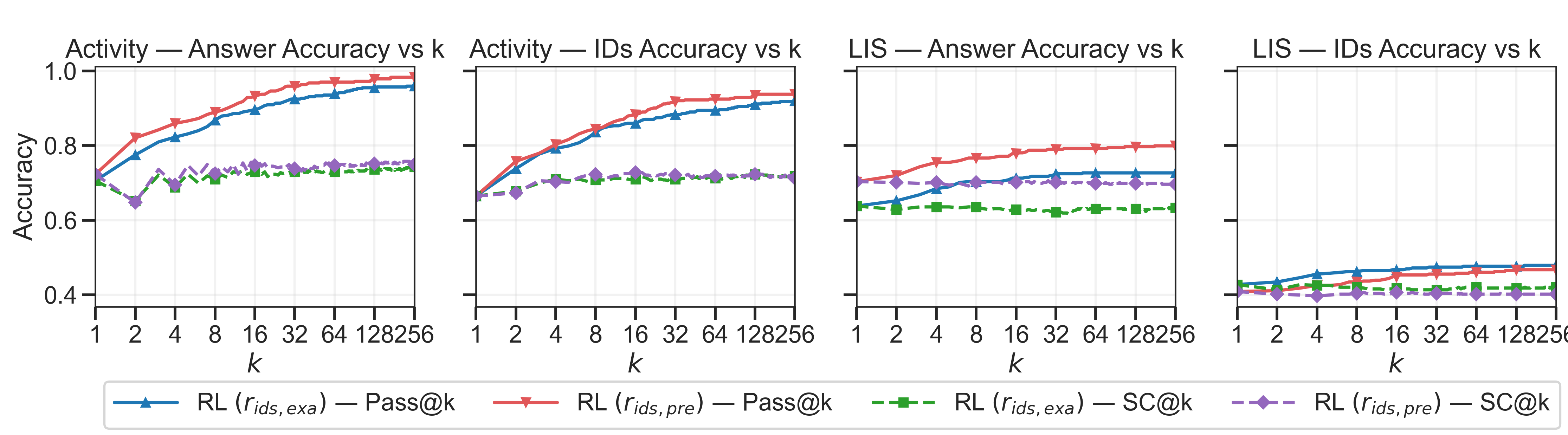}

\caption{Performance comparison of RL models trained with $r_{ids,exa}$ and $r_{ids,pre}$.}

    \label{fig:actlis_ids}

\end{figure}





\subsection{Training with Sequence Rewards}
\label{sec:seqreward}
We now evaluate sequence-aware objectives, comparing the exact-match reward $r_{\text{ids,exa}}$ and the prefix reward $r_{\text{ids,pre}}$ (cf.\ §\ref{sec:rewards}). Figure~\ref{fig:actlis_ids} summarizes results for $\mathrm{Acc}_{\text{ans}}$ and $\mathrm{Acc}_{\text{ids}}$ across $k$. On \textit{Activity}, both rewards yield similar gains, substantially surpassing the base and $r_{\text{ans}}$ models: at $k{=}256$, $\mathrm{SC}_{\text{ids}}$ rises from $0.34$ to $0.72/0.71$ ($r_{\text{ids,exa}}$/$r_{\text{ids,pre}}$), and $\mathrm{SC}_{\text{ans}}$ increases from $0.68$ to $0.74/0.75$. On \textit{LIS}, a trade-off emerges: $r_{\text{ids,pre}}$ attains higher $\mathrm{Acc}_{\text{ans}}$, while $r_{\text{ids,exa}}$ achieves higher $\mathrm{Acc}_{\text{ids}}$, yet both outperform the base and $r_{\text{ans}}$ policies. At $k{=}256$, $\mathrm{SC}_{\text{ids}}$ improves from $\approx 0.08$ to $0.42/0.40$, and $\mathrm{SC}_{\text{ans}}$ from $0.58$ to $0.63/0.70$, indicating that sequence rewards enhance both answer- and sequence-level consistency.


\begin{tcolorbox}[colback=gray!10, colframe=black, title=Takeaway 2]
Sequence-aware rewards improve sequence fidelity on both \textit{Activity Scheduling} and \textit{LIS}, as reflected in higher $\mathrm{Acc}_{\text{ids}}$. 
They also provide modest secondary gains in $\mathrm{Acc}_{\text{ans}}$, suggesting that RLVR generalization can benefit from incorporating intermediate or auxiliary objectives.
\end{tcolorbox}

\subsection{Sorting Performance on Activity Task}
\label{sec:sorting_performance}

Sequence-aware rewards improve sequence matching partly by enhancing the “sorting preface” that models generate as the first step in activity scheduling. We extract candidate ID lists from outputs using simple patterns, succeeding on over $84\%$ of RL-trained responses versus $40\%$ for the base model (Appendix~\ref{app:sorted-extraction}). Evaluating \textbf{ExactSort} (exact match with ground truth) and \textbf{LCS fraction} (longest correctly sorted subsequence) shows that sequence rewards yield both higher exact sorting and better partial order fidelity.

\begin{figure}[t]
  \centering
  \begin{subfigure}[t]{0.48\linewidth}
    \centering
    \includegraphics[width=\linewidth]{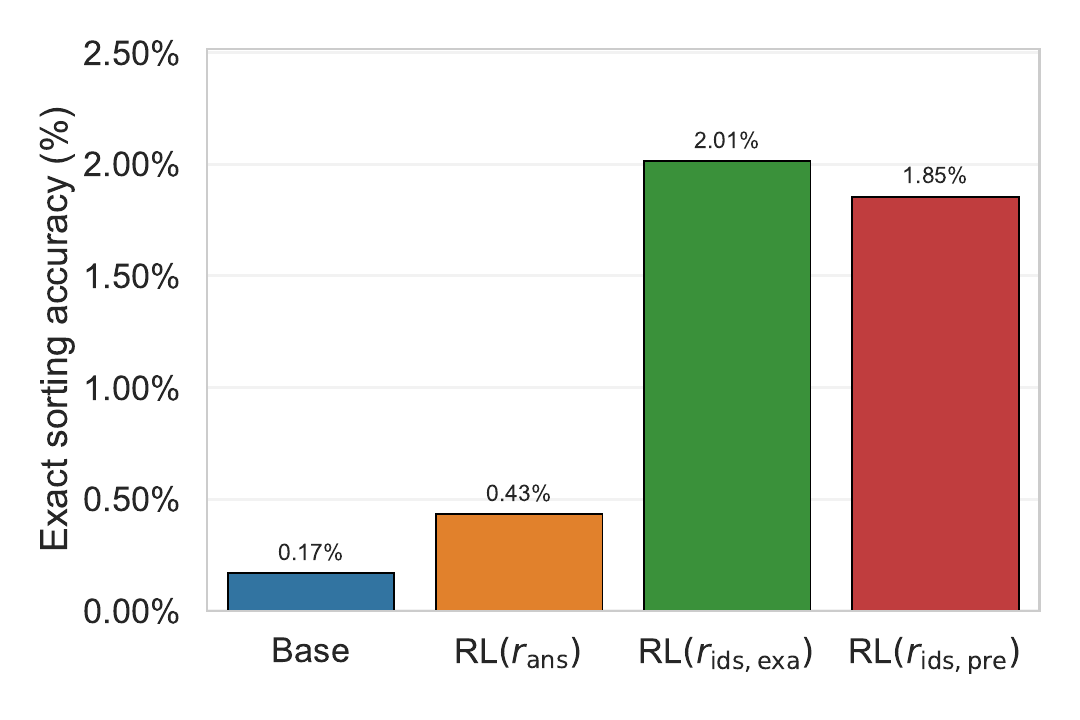}
    
    \label{fig:sorting-acc-bar}
  \end{subfigure}\hfill
  \begin{subfigure}[t]{0.48\linewidth}
    \centering
    \includegraphics[width=\linewidth]{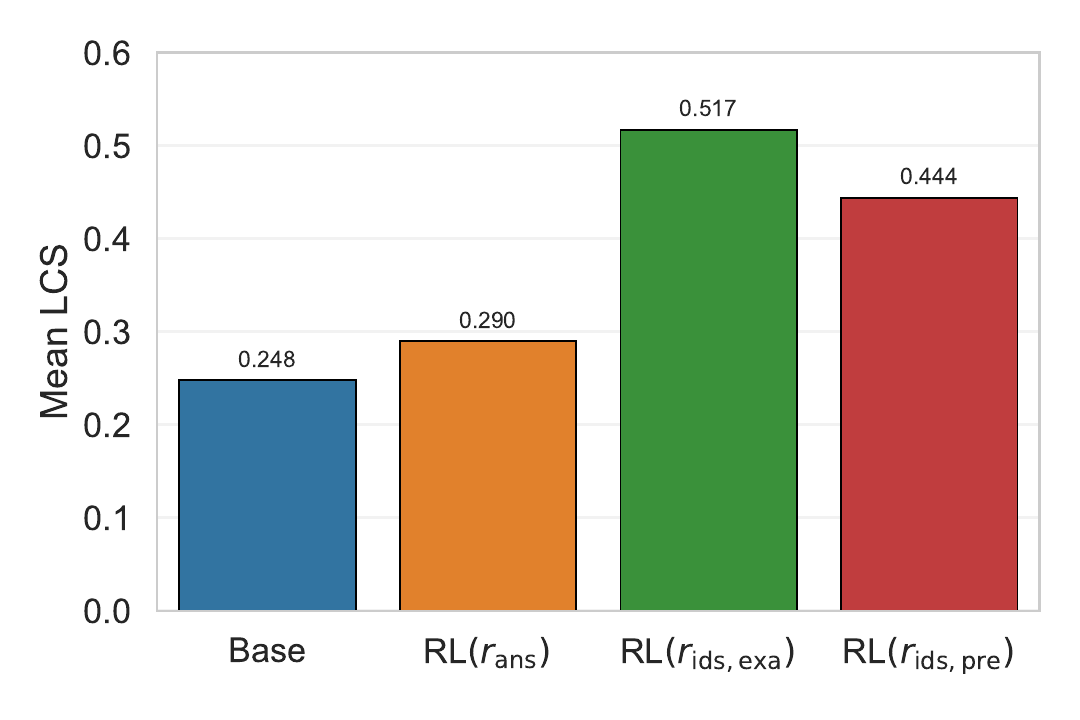}
    
    \label{fig:mean-lcs-bar}
  \end{subfigure}
  \caption{Sorting accuracy and LCS across models.}
  \label{fig:sorting-lcs-two-bars}
\end{figure}

Figure~\ref{fig:sorting-lcs-two-bars} summarizes sorting quality for the four models. 
Both sequence-reward policies ($r_{\text{ids,exa}}$ and $r_{\text{ids,pre}}$) improve \emph{exact sorting} relative to the base and $r_{\text{ans}}$ models (from $0.17\%/0.43\%$ to $2.01\%/1.85\%$) and increase the mean LCS (from $0.248/0.290$ to $0.517/0.444$). 
However, absolute exact-sorting accuracy remains very low ($\approx\!2\%$), even though these same policies achieve high sequence correctness at evaluation (e.g., $\mathrm{Pass@256}_{\text{ids}}\!\approx\!0.60$ on \textit{Activity}). 
This gap indicates that the sorting step is not reliably driving the final schedule, despite the frequent appearance of a “sorted” preface.

\subsection{Training with Sorting-Match Reward}
\label{sec:sortreward}

In \S\ref{sec:seqreward}, we observed performance gains in $\mathrm{Acc}_{\text{ans}}$ when training with sequence-aware rewards. This raises the question of whether rewarding the sorting step, which is the first stage of the greedy activity-scheduling algorithm, could provide additional benefit. To test this, we trained a model with the sorting-match reward $r_{\text{sort}}$. 

Surprisingly, this led to a catastrophic collapse: both $\mathrm{Acc}_{\text{ans}}$ and $\mathrm{Acc}_{\text{ids}}$ fell to nearly $0\%$. Inspection of the outputs revealed that the model consistently produced the sorted sequence itself as the final answer, without applying the non-overlap constraint required for valid schedules. 

We next trained a model using a combined objective with equal weights on $r_{\text{ans}}$, $r_{\text{ids}}$, and $r_{\text{sort}}$. 
This configuration restored $\mathrm{Acc}_{\text{ans}}$ and $\mathrm{Acc}_{\text{ids}}$ to levels comparable to the $r_{\text{ids}}$ model, but it did not improve sorting performance on the test set. 
Thus, learning to sort in isolation does not confer complementary benefits for solving the activity-scheduling task. 

\begin{figure}[t]
  \centering
  \begin{subfigure}[t]{0.48\linewidth}
    \centering
    \includegraphics[width=\linewidth]{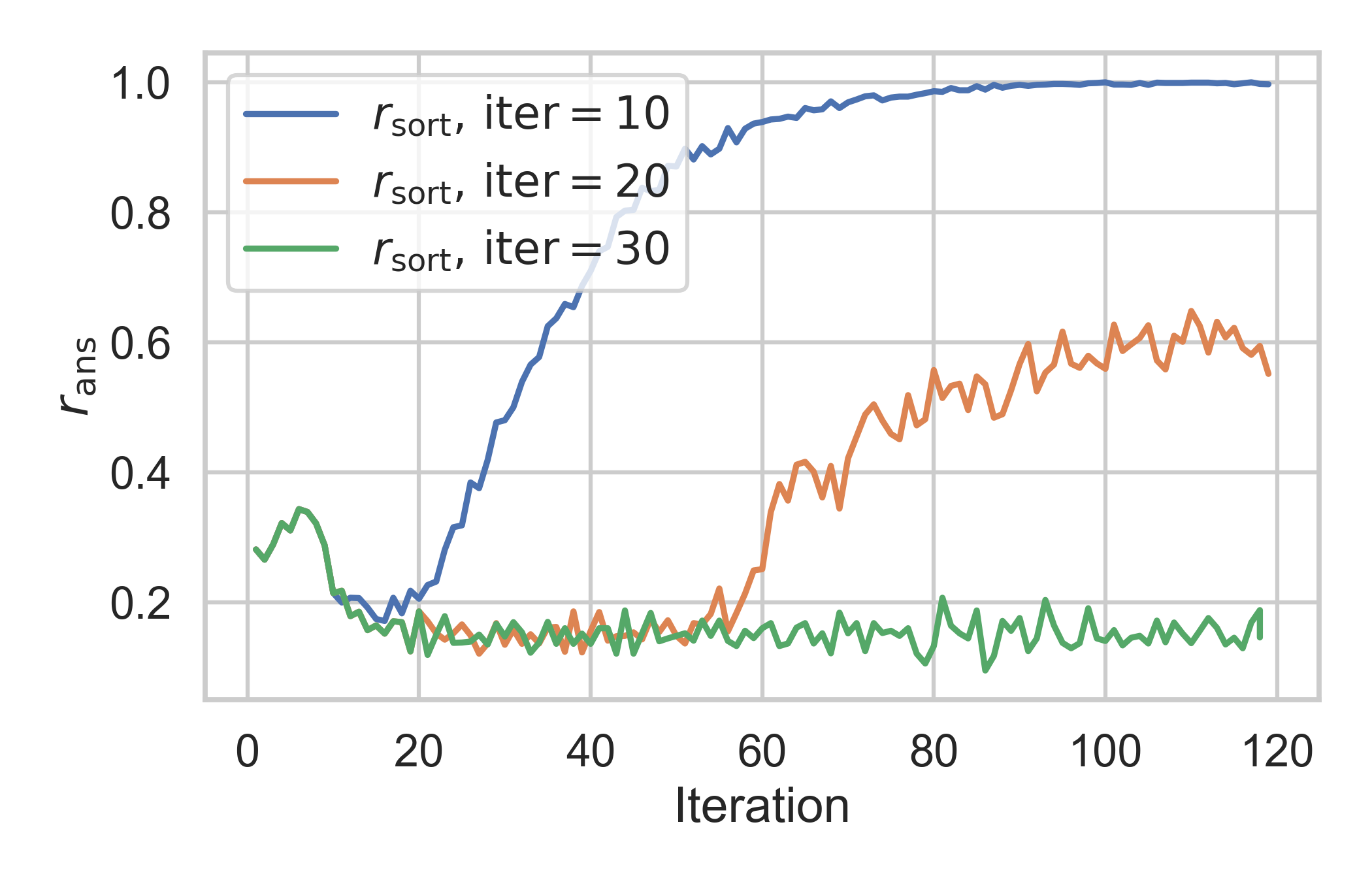}
    \label{fig:sorting-acc-bar}
  \end{subfigure}\hfill
  \begin{subfigure}[t]{0.48\linewidth}
    \centering
    \includegraphics[width=\linewidth]{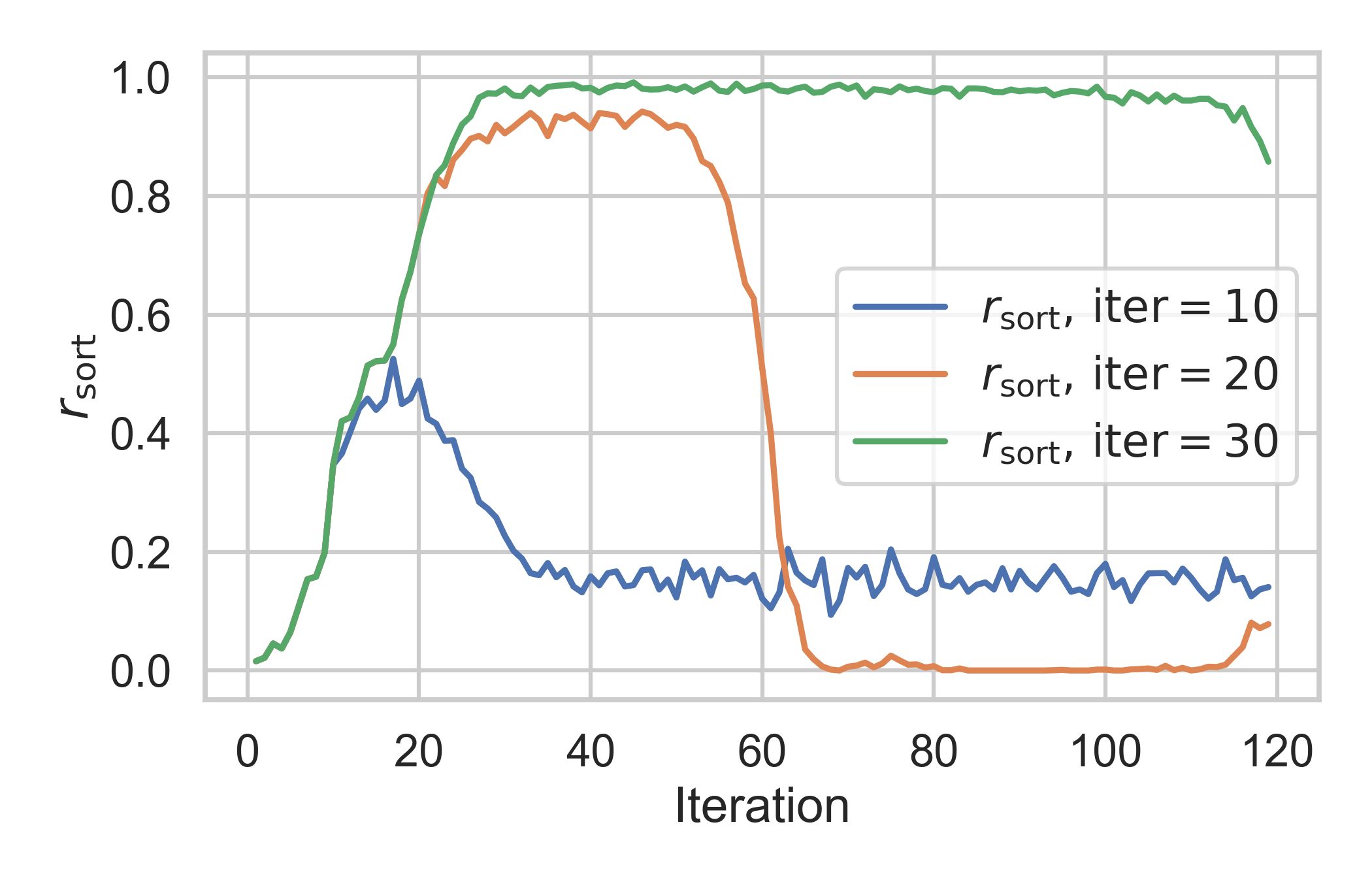}
    \label{fig:mean-lcs-bar}
  \end{subfigure}
  \caption{Curriculum experiments with $r_{\text{sort}}$. 
  Each curve shows accuracy on the training set when models are trained with $r_{\text{sort}}$ for the first 10, 20, or 30 PPO updates, followed by $r_{\text{ans}}$ for the remainder. 
  Longer pretraining with $r_{\text{sort}}$ makes it increasingly difficult for the model to recover under $r_{\text{ans}}$.}
  \label{fig:acc_curiculym}
\end{figure}

To probe further, we ran a curriculum experiment: training with $r_{\text{sort}}$ alone for the first 10, 20, or 30 PPO updates, and then switching to $r_{\text{ans}}$ for the remainder of the 120 training steps. 
As shown in Figure~\ref{fig:acc_curiculym}, models trained with $r_{\text{sort}}$ for only 10 steps were able to recover under $r_{\text{ans}}$, but longer exposure (20 or 30 steps) severely hindered recovery. 
In fact, after 30 steps of $r_{\text{sort}}$, the model failed to improve $\mathrm{Acc}_{\text{ans}}$ even on the training data.

\begin{tcolorbox}[colback=gray!10, colframe=black, title=Takeaway 3]
In \textit{Activity Scheduling}, RLVR with sequence rewards improves answer and sequence accuracy, but sorting accuracy remains very low. Models often emit a superficial “sorted” preface that neither matches the canonical order nor drives the final schedule, highlighting a disconnect between surface outputs and the underlying decision rule~\cite{turpin2023language}.
\end{tcolorbox}

\subsection{Heuristics Analysis for LIS}
To probe what signals models exploit on \textit{LIS}, we regress predicted numeric answers against human-interpretable features from the input table. 

\begin{wraptable}{r}{0.31\linewidth}
\centering
\small
\begin{tabular}{lcc}
\toprule
Model & $R^2_{\text{test}}$ & $\text{MAE}_{\text{test}}$ \\
\midrule
Base & -0.002 & 2.745 \\
$r_{\text{ans}}$ & 0.741 & 0.269 \\
$r_{\text{ids,exa}}$ & 0.781 & 0.289 \\
$r_{\text{ids,pre}}$ & 0.841 & 0.227 \\
$r_{\text{ans+fmt}}$ & 0.867 & 0.209 \\
\bottomrule
\end{tabular}
\caption{Predictive fit of LIS\\ features on model outputs.}
\label{tab:lis_r2_mae}
\vspace{-0.8em}
\end{wraptable}

We pool all $k$ stochastic runs but split train/test by problem instance ID to avoid leakage. 
Features cover: (i) global scale (length, range, dispersion, quantiles), (ii) order structure (increase ratios, inversion ratio, sign changes), (iii) run structure (longest runs, monotone counts, local extrema, record highs/lows), (iv) simple LIS heuristics (greedy, beam-limited, limited backtracking), and (v) patience-sorting tails. 
We train a Random Forest regressor and report $R^2$ and mean absolute error (MAE). Details appear in Appendix~\ref{app:rf}.

Table~\ref{tab:lis_r2_mae} shows that RLVR-trained outputs are predicted well from these features, unlike the base model. 
While the features are not exhaustive, results suggest RLVR amplifies systematic heuristics aligned with task structure, boosting answer-level performance without ensuring stronger reasoning.
\section{Limitations \& Future Work}
Our study investigates two verifiable reasoning tasks (\textit{Activity Scheduling} and \textit{LIS}) with a single base model (\textit{Qwen2.5-7B-Instruct}). While this controlled setup helps isolate RLVR dynamics, conclusions may not generalize across tasks, model families, or scales, as prior work has shown diverse shortcut behaviors and model-dependent failure modes~\cite{shao2025spurious}. To partially address this, we provide additional results for \textit{Llama-3.1-8B} in Appendix~\ref{sec:llama3}.  

Future work could expand to a broader range of models and problem domains, and employ mechanistic interpretability to probe the circuits and dynamics underlying RLVR-induced behaviors~\cite{sharkey2025open}. Such analysis may clarify whether improvements reflect genuine task learning or superficial strategies that exploit evaluation metrics. Our findings highlight this tension, emphasizing the need for careful evaluation and diagnostics when claiming improvements in reasoning from RLVR.

\clearpage

\newpage

\bibliographystyle{ACM-Reference-Format}
\bibliography{references}


\appendix

\clearpage
\section*{Appendix}

\section{Dataset Construction}
\label{app:data-generator}

\subsection{Activity Scheduling}
\label{app:activity-generator}

\paragraph{Problem model.}
We work on the standard unweighted interval scheduling problem.
Each instance has $m$ activities indexed by $I=\{1,\dots,m\}$ with start/finish times $(s_i,f_i)$ and $s_i<f_i$.
Intervals are treated as \emph{half-open} $[s_i,f_i)$, so touching endpoints are compatible ($f_j\le s_i$ means $i$ and $j$ can both be selected).
We fix a deterministic tie-breaking order by the tuple $(f_i,s_i,i)$ whenever sorting is required.
The canonical reporting order for ground truth is by non-decreasing $f$ (ties by smaller $i$).

\paragraph{Sampling on a minute grid.}
All times lie on an integer minute grid.
To construct a candidate set of $m$ activities, we:
(i) sample $m\sim\mathrm{Unif}\{m_{\min},\dots,m_{\max}\}$;
(ii) repeatedly sample integer start times $s\sim\mathrm{Unif}\{0,\dots,S_{\max}\}$ and integer durations $d\sim\mathrm{Unif}\{D_{\min},\dots,D_{\max}\}$, set $f=s+d$, and accept the interval if $f\le S_{\max}+D_{\max}$; we continue until $m$ intervals are accepted.
(Values used in our experiments: $m_{\min}=5$, $m_{\max}=16$, $S_{\max}=9\!\times\!60$, $D_{\min}=10$, $D_{\max}=120$; any fixed choices are acceptable as long as $D_{\min}\ge 1$ and times are integral.)
Each accepted interval receives a stable ID $i\in\{1,\dots,m\}$ in order of creation; the table shown to the model lists rows by ID with times in \texttt{HH:MM}.

\paragraph{Uniqueness check by DP counting.}
Write $J=(1,\dots,n)$ for the intervals \emph{sorted} by $(f_i,s_i,i)$ (we reuse the symbol $i$ for the sorted index when clear).
Define the predecessor map
\[
p(i)\;=\;\max\{\, j<i \;:\; f_j \le s_i \,\}\quad\text{with }p(i)=0\text{ if the set is empty.}
\]
Let $\mathrm{opt}[i]$ be the maximum feasible cardinality using only $\{1,\dots,i\}$ and let $\mathrm{cnt}[i]$ be the \emph{number of distinct schedules} that attain $\mathrm{opt}[i]$ using $\{1,\dots,i\}$.
Initialize $\mathrm{opt}[0]=0$ and $\mathrm{cnt}[0]=1$ (one empty schedule).
For $i=1,\dots,n$ set
\[
\begin{aligned}
\textsc{incl} &= 1 + \mathrm{opt}[\,p(i)\,],\\
\textsc{excl} &= \mathrm{opt}[\,i-1\,],\\[2pt]
\mathrm{opt}[i] &= \max\{\textsc{incl},\textsc{excl}\},\\
\mathrm{cnt}[i] &=
\begin{cases}
\mathrm{cnt}[\,p(i)\,], & \text{if }\textsc{incl}>\textsc{excl},\\
\mathrm{cnt}[\,i-1\,],   & \text{if }\textsc{excl}>\textsc{incl},\\
\mathrm{cnt}[\,p(i)\,]+\mathrm{cnt}[\,i-1\,], & \text{if }\textsc{incl}=\textsc{excl}.
\end{cases}
\end{aligned}
\]
Then $k^\star=\mathrm{opt}[n]$ is the optimal size and $\mathrm{cnt}[n]$ is the number of optimal schedules.
If $\mathrm{cnt}[n]=1$, the optimum is unique.
Under uniqueness we reconstruct the unique optimal set $S^\star$ by backtracking from $i=n$:
when $\textsc{incl}>\textsc{excl}$ we \emph{include} $i$ and jump to $p(i)$; when $\textsc{excl}>\textsc{incl}$ we \emph{exclude} $i$ and go to $i-1$; when equal, exactly one branch leads to a nonzero count under uniqueness—choose the branch whose downstream count equals $1$ (include if $\mathrm{cnt}[p(i)]=1$ and $\mathrm{cnt}[i-1]=0$, otherwise exclude).

\paragraph{Greedy sanity check and canonicalization.}
Because earliest-finish-time is optimal for unweighted interval scheduling, if the optimum is unique then the greedy policy with the same tie-breaks must return that same set.
We nevertheless perform a \emph{sanity check}: run a single pass of earliest-finish-time using the order $(f_i,s_i,i)$ to obtain $G$; we accept the instance only if $G=S^\star$ (as sets).
The ground-truth sequence we release is the unique set $S^\star$ \emph{sorted by non-decreasing $f$ (ties by $i$)} and the ground-truth answer is $|S^\star|$.
This yields a single canonical \verb|\ids{...}| and \verb|\answer{...}| for each instance.

\begin{algorithm}[t]
\caption{\textsc{CountOptimaAndBacktrack} for Interval Scheduling}
\label{alg:count-unique}
\KwIn{Intervals with IDs $(i,s_i,f_i)$, $i=1,\dots,m$.}
\KwOut{Optimal size $k^\star$, number of optimal schedules $\mathrm{cnt}[n]$, unique set $S^\star$ if $\mathrm{cnt}[n]=1$.}
Sort intervals by $(f_i,s_i,i)$ to obtain $J=(1,\dots,n)$; precompute $p(i)=\max\{j<i:\ f_j\le s_i\}$ by binary search\;
$\mathrm{opt}[0]\gets 0$, $\mathrm{cnt}[0]\gets 1$\;
\For{$i\gets 1$ \KwTo $n$}{
  $\textsc{incl}\gets 1+\mathrm{opt}[p(i)]$, $\textsc{excl}\gets \mathrm{opt}[i-1]$\;
  $\mathrm{opt}[i]\gets \max\{\textsc{incl},\textsc{excl}\}$\;
  \lIf{$\textsc{incl}>\textsc{excl}$}{$\mathrm{cnt}[i]\gets \mathrm{cnt}[p(i)]$}
  \lElseIf{$\textsc{excl}>\textsc{incl}$}{$\mathrm{cnt}[i]\gets \mathrm{cnt}[i-1]$}
  \lElse{$\mathrm{cnt}[i]\gets \mathrm{cnt}[p(i)]+\mathrm{cnt}[i-1]$}
}
$k^\star\gets \mathrm{opt}[n]$\;
\lIf{$\mathrm{cnt}[n]\neq 1$}{\Return{$(k^\star,\,\mathrm{cnt}[n],\,\emptyset)$}}
\tcp{Uniqueness holds: recover the single optimal set}
$S^\star\gets\emptyset$; $i\gets n$\;
\While{$i>0$}{
  $\textsc{incl}\gets 1+\mathrm{opt}[p(i)]$, $\textsc{excl}\gets \mathrm{opt}[i-1]$\;
  \uIf{$\textsc{incl}>\textsc{excl}$}{add $i$ to $S^\star$; $i\gets p(i)$}
  \uElseIf{$\textsc{excl}>\textsc{incl}$}{$i\gets i-1$}
  \Else{
    \uIf{$\mathrm{cnt}[p(i)]=1$ and $\mathrm{cnt}[i-1]=0$}{add $i$ to $S^\star$; $i\gets p(i)$}
    \Else{$i\gets i-1$}
  }
}
\Return{$(k^\star,\,1,\,S^\star)$}\tcp*[f]{Sort $S^\star$ by $(f,i)$ when reporting.}
\end{algorithm}

\begin{algorithm}[t]
\caption{\textsc{GenerateUniqueActivityInstance}}
\label{alg:gen-unique-activity}
\KwIn{RNG seed; $m_{\min},m_{\max}$; minute-grid bounds $S_{\max}$, $D_{\min},D_{\max}$; $\texttt{max\_tries}$.}
\KwOut{Intervals $\{(i,s_i,f_i)\}_{i=1}^m$; canonical \texttt{\textbackslash ids\{\}} sequence; integer answer $|S^\star|$.}

\For{$t\gets 1$ \KwTo $\texttt{max\_tries}$}{
  Sample $m\sim\mathrm{Unif}\{m_{\min},\dots,m_{\max}\}$\;
  $I\gets\emptyset$\;
  \While{$|I|<m$}{
    sample $s\sim\mathrm{Unif}\{0,\dots,S_{\max}\}$, $d\sim\mathrm{Unif}\{D_{\min},\dots,D_{\max}\}$, set $f\gets s+d$\;
    \lIf{$f\le S_{\max}+D_{\max}$}{append new interval $(|I|{+}1,s,f)$ to $I$}
  }
  $(k^\star,\mathrm{cnt},S^\star)\gets\textsc{CountOptimaAndBacktrack}(I)$\;
  \lIf{$\mathrm{cnt}\neq 1$}{\textbf{continue}}
  $G\gets$ greedy earliest-finish schedule on $I$ using $(f,s,i)$ tie-breaks\;
  \lIf{$G\neq S^\star$}{\textbf{continue}}
  \tcp{Canonical target: IDs of $S^\star$ sorted by $(f,i)$; answer $=|S^\star|$}
  \Return{$(I,\ \texttt{\textbackslash ids}\{\,\text{IDs}(S^\star)\,\},\ \texttt{\textbackslash answer}\{\,|S^\star|\,\})$}
}
\textbf{Fail} if no instance is found within $\texttt{max\_tries}$ (try another seed or bounds)\;
\end{algorithm}

\paragraph{Complexity.}
Preprocessing and $p(i)$ by binary search take $O(m\log m)$; the DP pass and backtracking are $O(m)$; greedy verification is $O(m)$.

\subsection{Longest Increasing Subsequence (Unique Optimum)}
\label{app:lis-generator}

\paragraph{Problem model.}
Given integers $a_1,\ldots,a_n$, an LIS is a subsequence
$1 \le i_1 < \cdots < i_L \le n$ with $a_{i_1} < \cdots < a_{i_L}$.
We allow duplicate values in $(a_i)$; the strict inequality enforces strictly increasing values.
The canonical target we release per instance is:
(i) the unique LIS \emph{index} sequence listed in increasing row order as
\verb|\ids{...}|, and (ii) its length $L$ as \verb|\answer{L}|.  \emph{IDs are 1-based row indices}.

\paragraph{Sampling.}
We work on an integer grid. For each trial we:
(i) draw a length $m \sim \mathrm{Unif}\{m_{\min},\dots,m_{\max}\}$;
(ii) draw values i.i.d. $a_i \sim \mathrm{Unif}\{V_{\min},\dots,V_{\max}\}$;
(iii) accept the instance only if the LIS is \emph{unique} (by index sequence) and $L \ge 2$.
Typical bounds used in our experiments are $m_{\min}{=}5$, $m_{\max}{=}16$, $V_{\min}{=}1$, $V_{\max}{=}1000$, but any fixed choices are valid.

\paragraph{Uniqueness check by $O(n^2)$ counting.}
Let $\texttt{len\_end}[i]$ be the LIS length ending at $i$, and $\texttt{cnt\_end}[i]$ the number of LIS that end at $i$ with that maximum length. Initialize $\texttt{len\_end}[i]{=}1$, $\texttt{cnt\_end}[i]{=}1$. For each $i=1..n$ and $j<i$:
\[
\text{if } a_j<a_i:
\quad
\begin{cases}
\text{if } \texttt{len\_end}[j]{+}1 > \texttt{len\_end}[i]: &
\texttt{len\_end}[i] \leftarrow \texttt{len\_end}[j]{+}1,\;
\texttt{cnt\_end}[i] \leftarrow \texttt{cnt\_end}[j] \\
\text{else if } \texttt{len\_end}[j]{+}1 = \texttt{len\_end}[i]: &
\texttt{cnt\_end}[i] \leftarrow \texttt{cnt\_end}[i] + \texttt{cnt\_end}[j].
\end{cases}
\]
Let $L=\max_i \texttt{len\_end}[i]$ and $\#\mathrm{LIS}=\sum_{i: \texttt{len\_end}[i]=L} \texttt{cnt\_end}[i]$.
We declare \emph{uniqueness} iff $\#\mathrm{LIS}=1$.

\paragraph{Canonical reconstruction (used only after uniqueness holds).}
We reconstruct one LIS index sequence via patience sorting with predecessor links:
scan left to right; for each $a_i$ place it by lower\_bound (first position $\ge a_i$) in the tails array;
store for each $i$ a predecessor pointer to the last index at the previous length; then backtrack from the final tail index to obtain indices in increasing order. 

\begin{algorithm}[t]
\caption{\textsc{CountLisLengthAndNumber} (strict LIS length \& count)}
\label{alg:lis-count}
\KwIn{Sequence $(a_1,\ldots,a_n) \in \mathbb{Z}^n$.}
\KwOut{$L$ (LIS length), $\#\mathrm{LIS}$ (number of LIS).}
\For{$i \gets 1$ \KwTo $n$}{ $\texttt{len\_end}[i]\gets 1$;\; $\texttt{cnt\_end}[i]\gets 1$ }
\For{$i \gets 1$ \KwTo $n$}{
  \For{$j \gets 1$ \KwTo $i{-}1$}{
    \If{$a_j < a_i$}{
      \uIf{$\texttt{len\_end}[j]{+}1 > \texttt{len\_end}[i]$}{
        $\texttt{len\_end}[i] \gets \texttt{len\_end}[j]{+}1$;\;
        $\texttt{cnt\_end}[i] \gets \texttt{cnt\_end}[j]$}
      \uElseIf{$\texttt{len\_end}[j]{+}1 = \texttt{len\_end}[i]$}{
        $\texttt{cnt\_end}[i] \gets \texttt{cnt\_end}[i] + \texttt{cnt\_end}[j]$}
    }
  }
}
$L \gets \max_i \texttt{len\_end}[i]$;\quad
$\#\mathrm{LIS} \gets \sum_{i:\texttt{len\_end}[i]=L}\texttt{cnt\_end}[i]$\;
\Return{$(L,\#\mathrm{LIS})$} \tcp*[f]{Runtime $O(n^2)$}
\end{algorithm}

\begin{algorithm}[t]
\caption{\textsc{GenerateUniqueLISInstance}}
\label{alg:gen-unique-lis}
\KwIn{RNG seed; $m_{\min},m_{\max}$; value bounds $V_{\min},V_{\max}$; $\texttt{max\_tries}$.}
\KwOut{Values $(a_1,\ldots,a_m)$; canonical \texttt{\textbackslash ids$\{\cdot\}$}; \texttt{\textbackslash answer$\{\cdot\}$}.}
\For{$t \gets 1$ \KwTo $\texttt{max\_tries}$}{
  Sample $m \sim \mathrm{Unif}\{m_{\min},\dots,m_{\max}\}$;\quad
  Sample $a_i \sim \mathrm{Unif}\{V_{\min},\dots,V_{\max}\}$ i.i.d.\ for $i{=}1..m$\;
  $(L,\#)\gets \textsc{CountLisLengthAndNumber}(a_{1:m})$\;
  \lIf{$L<2$ \textbf{ or } $\# \neq 1$}{\textbf{continue}}
  \tcp{Reconstruct unique LIS indices via patience sorting with predecessors}
  $(i_1,\ldots,i_L)\gets \textsc{PatienceReconstruct}(a_{1:m})$ 
  \Return{$\big(a_{1:m},\,\texttt{\textbackslash ids}\{i_1,\ldots,i_L\},\,\texttt{\textbackslash answer}\{L\}\big)$}
}
\textbf{Fail} if no unique instance within $\texttt{max\_tries}$ (try new seed or bounds)\;
\end{algorithm}

\paragraph{Canonical reporting and prompts.}
Because the LIS is a subsequence, the ground-truth \verb|\ids{...}| is simply the unique LIS indices
$(i_1<\cdots<i_L)$ in increasing row order; \verb|\answer{L}| reports its length. 

\paragraph{Complexity.}
Per trial: $O(n^2)$ for counting uniqueness; $O(n\log n)$ to reconstruct the indices once unique.
Overall generation is bounded by \texttt{max\_tries}, which we set large enough so failures are rare.

\setlist[itemize]{leftmargin=1.2em, topsep=0.2em, itemsep=0.1em} 
\setlist[enumerate]{leftmargin=1.6em, topsep=0.2em, itemsep=0.1em} 
\begin{figure}[t] 
\centering 
\setlength{\tabcolsep}{6pt} 
\renewcommand{\arraystretch}{1.0} 
\begin{tabular}{@{}p{0.485\linewidth} p{0.485\linewidth}@{}}
\begin{tcolorbox}[colback=white,colframe=black!60,boxrule=0.5pt, sharp corners,enhanced jigsaw, equal height group=A, title=Activity Scheduling Prompt] 
\scriptsize 
Determine the largest subset of activities that can be scheduled without any overlaps (a single resource is available, so no double-booking). \\
\vspace{0.4em} 
\begin{tabular}{@{}rcc@{}} 
ID & Start & End\\ \hline 
1 & 06:09 & 07:24 \\ 
2 & 07:13 & 08:23 \\ 
3 & 07:29 & 09:28 \\ 
4 & 08:24 & 10:18 \\ 
5 & 04:48 & 06:14 \\ 
\end{tabular} 
\vspace{0.4em} \\
Instructions: 
\begin{itemize} 
\item Times are given in 24-hour HH:MM format. 
\item Non-overlap means an activity ending at time $T$ is compatible with one starting at $T$.
\item Select a maximum-size subset of non-overlapping rows.
\item Your output must include the following two lines at the end, in this exact format:
\begin{enumerate}
\item \texttt{\string\ids\{<comma-separated IDs of the chosen rows, listed in order of increasing END time. If two rows have the same END time, put the smaller ID first>\}}
\item \texttt{\string\answer\{<number of chosen rows>\}}
\end{enumerate} 
\item No spaces inside \texttt{\string\ids\{...\}}. Example: \texttt{\string\ids\{3,9,12\}}
\end{itemize} 
Hint: Sort the rows by increasing end time, then greedily pick compatible rows. \\
\vspace{0.3em}  \\
\textbf{Ground truth:} \texttt{\string\ids\{5,2,4\}}, \texttt{\string\answer\{3\}} 
\end{tcolorbox} & 
\begin{tcolorbox}[colback=white,colframe=black!60,boxrule=0.5pt, sharp corners,enhanced jigsaw, equal height group=A, title=LIS Prompt] 
\scriptsize 
Determine the longest strictly increasing subsequence (by VALUE) from the rows below (use the row IDs). \\
\vspace{0.4em} 
\begin{tabular}{@{}rc@{}} 
ID & Value\\ \hline 
1 & 797 \\ 
2 & 476 \\ 
3 & 335 \\ 
4 & 452 \\ 
5 & 606 \\ 
\end{tabular} 
\vspace{0.4em} \\
Instructions: 
\begin{itemize} 
\item The table lists 1-based row IDs and integer values.
\item A valid subsequence must be strictly increasing by VALUE and preserve the original row order.
\item Select a maximum-size subsequence.
\item Your output must include the following two lines at the end, in this exact format:
\begin{enumerate}
\item \texttt{\string\ids\{<comma-separated IDs of the chosen rows, listed in increasing ROW order>\}}
\item \texttt{\string\answer\{<number of chosen rows>\}}
\end{enumerate} 
\item No spaces inside \texttt{\string\ids\{...\}}. Example: \texttt{\string\ids\{3,9,12\}}
\end{itemize}
\vspace{1.5em}
Hint: Use LIS DP (len[i] = 1 + $\max$ len[j] for $j<i$ with value[j]<value[i]) while storing prev[i] (or patience sorting with predecessor links); then backtrack to output the ids.
\vspace{1.em} \\

\textbf{Ground truth:} \texttt{\string\ids\{3,4,5\}}, \texttt{\string\answer\{3\}} 
\end{tcolorbox} \\ 
\end{tabular} 
\caption{Example prompts and ground-truth for Activity Scheduling (left) and LIS (right).} 
\label{fig:prompt} 
\end{figure}

\section{Extracting sorted ID lists from free‑form responses}
\label{app:sorted-extraction}

\paragraph{Setup.}
For each Activity instance the prompt shows an ASCII table with unique integer IDs $i\in\{1,\dots,n\}$ and times $(s_i,f_i)$.  
We define the ground‑truth sorted order
\[
B=(b_1,\dots,b_n)\ :=\ \text{IDs sorted by }(f_i,i)\ \text{(non‑decreasing end time, ties by smaller ID)}.
\]
This $B$ is the order used by the greedy earliest‑finish‑time scheduler (Alg.~1, §3.1). From the model’s free‑form reply $r$ (arbitrary text) we attempt to recover a list the model claims to be “sorted.” 

\paragraph{Candidate sources (lightweight and order‑preserving).}
We build up to four candidate ID sequences from $r$; all candidates are \emph{normalized} by (i) filtering to the valid ID set $\mathcal{I}=\{1,\dots,n\}$ derived from the prompt table, and (ii) de‑duplicating while preserving the \emph{first} occurrence of each ID in the text.

\begin{enumerate}[leftmargin=1.5em,itemsep=2pt]
\item \textbf{Sorted‑block (for exact‑sorting only).} We split $r$ into paragraphs and select those that mention a sorting token (``sort/sorted/sorting''). To isolate the purported sorting step, we truncate at the first stop word (any of: \texttt{select, greedy, choose, subset, largest, final answer, so, thus, therefore, next}). We then extract IDs from this truncated text via either ``\texttt{ID $k$}'' tokens or the longest comma‑separated integer run. If the resulting list contains \emph{all} $n$ distinct IDs exactly once, we accept it as a full ``sorted‑block'' candidate $A_{\text{full}}$ and reserve it solely for the exact‑sorting check below.
\item \textbf{\texttt{\textbackslash ids\{...\}} blocks.} From every brace block we read the comma‑separated integers, normalize, and keep the resulting sequence if non‑empty.
\item \textbf{``ID $k$'' token stream.} We scan $r$ left‑to‑right for tokens of the form ``\texttt{ID $k$}’’ and record the first occurrence of each $k$.
\item \textbf{Longest comma run.} We find the longest comma‑separated integer run anywhere in $r$ and normalize it.
\end{enumerate}

These patterns capture the most common surface forms we observe in practice and are exactly those used in our implementation. 

\paragraph{Exact‑sorting criterion.}
If the sorted‑block candidate $A_{\text{full}}$ exists and is a permutation of all $n$ IDs (after normalization), we declare \emph{extraction success}. The instance counts as \emph{exactly sorted} iff $A_{\text{full}}=B$. Missing or malformed cases are treated as incorrect in the exact‑sorting accuracy.

\paragraph{Best‑of‑candidates policy and anchors (for substring analysis).}
For contiguous‑substring analysis (§\ref{app:lcs}) we consider \emph{all} non‑empty candidates
$\mathcal{A}=\{A_1,\dots,A_K\}$ formed by items 2–4 above (and $A_{\text{full}}$ if present).  
When multiple candidates tie on the score defined next, we break ties by method priority:
\[
\texttt{sorted\_block\_full}\ \prec\ \texttt{ids\_braces}\ \prec\ \texttt{id\_stream}\ \prec\ \texttt{comma\_run}.
\]
For the chosen best candidate we additionally report an \emph{anchor} label describing where the matched block sits inside the candidate: \texttt{start} (run begins at position~1), \texttt{end} (run ends at position $|A_k|$), \texttt{both} (the candidate equals the block), or \texttt{neither}.

\paragraph{Safeguards and edge cases.}
We ignore any integers not present in the prompt table; repeated IDs are dropped after the first mention; candidates that normalize to the empty list are discarded.  
These choices make the procedure robust to extraneous numbers and minor formatting quirks without conferring credit for unseen IDs. 

\paragraph{Complexity.}
Regex scans and candidate construction are $O(|r|)$; all subsequent scoring (§\ref{app:lcs}) is linear in the candidate length(s).

\section{Contiguous LCS (longest common substring) metric}
\label{app:lcs}

\paragraph{Definition.}
Given a candidate $A=(a_1,\dots,a_m)$ and the ground‑truth order $B=(b_1,\dots,b_n)$ (IDs unique), define the position map $\mathrm{pos}(b_\ell)=\ell$.  
Let $p_t=\mathrm{pos}(a_t)$ for $t=1,\dots,m$ (these indices exist by construction after normalization).  
The \emph{contiguous} LCS length between $A$ and $B$ is
\begin{equation}
\mathrm{LCS\_len}(A,B)
\;=\;
\max_{1\le i\le j\le m}\Big\{(j-i+1)\ \big|\ \exists\,t\ \text{s.t.}\ \forall\,u\in\{0,\dots,j-i\},\ p_{i+u}=t+u\Big\}.
\label{eq:contig-lcs}
\end{equation}
Intuitively, \eqref{eq:contig-lcs} is the length of the longest block in $A$ that appears \emph{contiguously} in $B$; because IDs in $B$ are unique, this is exactly the standard longest common \emph{substring}. 

\paragraph{Instance score and reporting.}
From the set of extracted candidates $\mathcal{A}=\{A_1,\dots,A_K\}$ we take the best match
\[
L^\star\;=\;\max_{k\in[K]}\ \mathrm{LCS\_len}(A_k,B),
\qquad
\mathrm{LCS\text{-}frac}\;=\;\frac{L^\star}{|B|}\;=\;\frac{L^\star}{n}.
\]
We report the mean LCS‑fraction over instances with $L^\star>0$ and a separate \emph{coverage} rate (\% of instances with any non‑empty match). Exact‑sorting accuracy and LCS are shown in Fig.~8. 

\paragraph{Anchors.}
For the candidate achieving $L^\star$ (after the tie‑break above), we also emit the anchor label \texttt{start/end/both/neither} based on whether the best contiguous block touches the candidate’s boundaries. Anchors help diagnose whether the model’s claimed “sorted” list appears as a leading or trailing segment versus a mid‑sequence fragment.

\section{LIS Task Response Length and Entropy}
\label{app:lismore}
\begin{figure}[t]
  \centering
  \begin{minipage}[t]{0.485\linewidth}
    \centering
    \includegraphics[width=\linewidth]{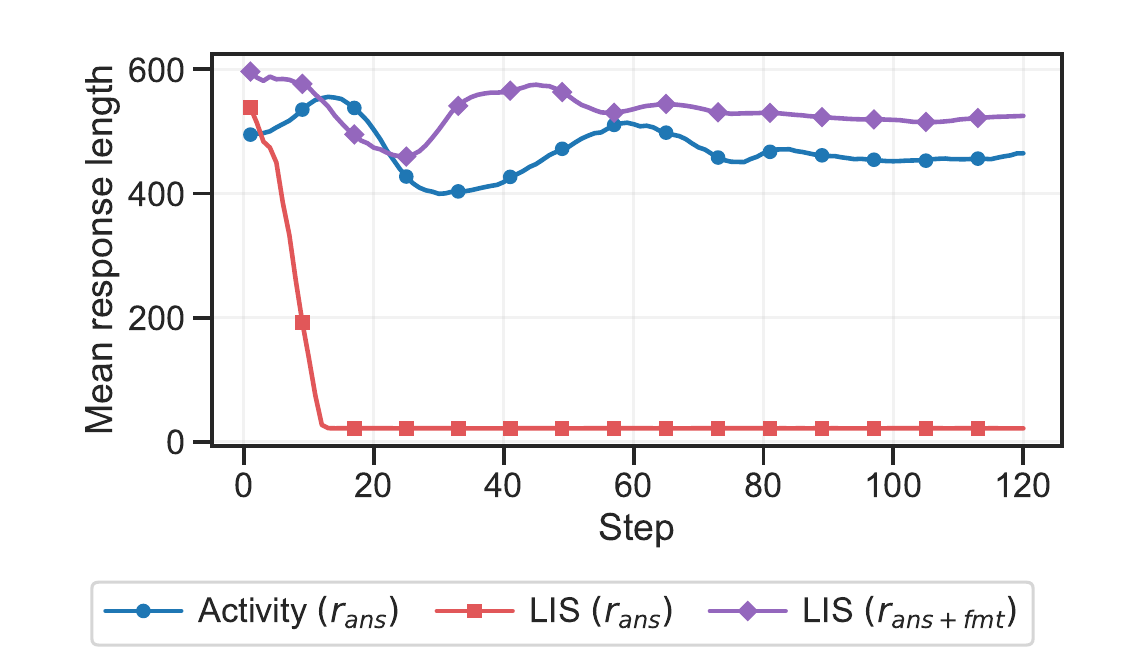}
    \captionof{figure}{Response length during training}
    \label{fig:response_len}
  \end{minipage}
  \hfill
  \begin{minipage}[t]{0.485\linewidth}
    \centering
    \includegraphics[width=\linewidth]{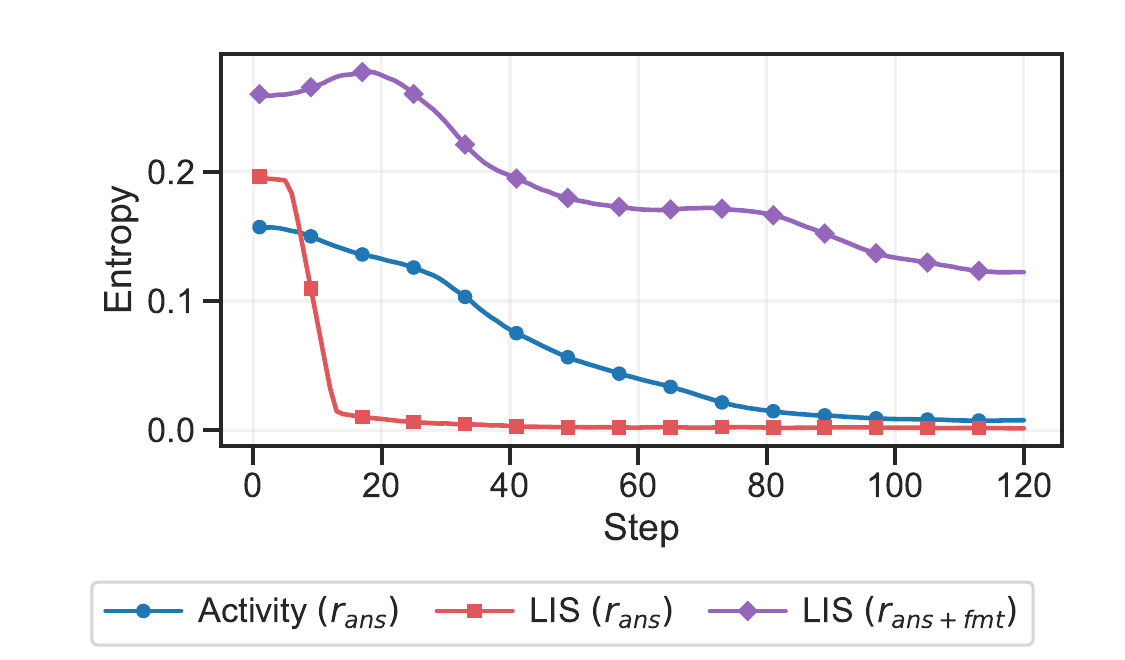}
    \captionof{figure}{Entropy during training}
    \label{fig:entropy}
  \end{minipage}
\end{figure}

Figure~\ref{fig:response_len} shows response length during training on the LIS task. Under the answer-only reward, responses quickly shorten, while the format reward preserves lengths close to those of the base model.

Figure~\ref{fig:entropy} further compares entropy during training. 
Entropy drops sharply for $r_{\text{ans}}$, mirroring the collapse in response length, whereas $r_{\text{ans+fmt}}$ maintains higher entropy throughout training. 
Despite these qualitative differences, both RLVR-trained policies converge to similar levels of $\mathrm{Acc}_{\text{ans}}$ and $\mathrm{Acc}_{\text{ids}}$, suggesting reliance on comparable heuristics expressed through different generation styles. 
For example, with $r_{\text{ans+fmt}}$ nearly all responses contain Python code (100\%, compared to 35.1\% for the base model), yet the model does not execute the code step by step to derive the answer. 
This observation echoes findings by~\citet{shao2025spurious}, who show that RLVR can amplify spurious reward signals in the Qwen family, encouraging models to emit structured but ultimately superficial outputs.

\section{Random-Forest Regression for LIS: Features, Protocol, and Training}
\label{app:rf}

\textbf{Goal.} We analyze the model’s implicit decision rule by regressing its numeric answer, $y=\texttt{pred\_lis\_len}$, on interpretable features computed \emph{only} from the input values $(v_1,\dots,v_n)$.

\paragraph{Data and target.}
From the JSONL logs we pool all stochastic runs $k$ for each instance (\texttt{sample\_idx}). The regression target is the model’s emitted \verb|\answer{\cdot}| value; no ground-truth labels are used as features.

\paragraph{Group split to avoid leakage.}
We perform a single group hold-out split with GroupShuffleSplit (test size $\approx 25\%$) using \texttt{sample\_idx} as the group key. Thus, all $k$ replicates of a given instance are kept together in train or test; there is \underline{no} overlap of \texttt{sample\_idx} between splits.

\paragraph{Feature set (input-only).}
Let $n$ be the sequence length and $\Delta_i = v_{i+1}-v_i$ for $i=1,\dots,n-1$. We compute:

\begin{itemize}
  \item \textbf{Global scale/dispersion:}
  $n$, $\min(v)$, $\max(v)$, $\mathrm{range}=\max-\min$, mean, std, quartiles $q_{25}, q_{50}, q_{75}$, 
  $\texttt{uniq\_ratio}=|\{v\}|/n$, and $\texttt{dup\_ratio}=1-\texttt{uniq\_ratio}$.
  \item \textbf{Adjacent order (local trend):}
  $\texttt{adj\_inc\_ratio}=\tfrac{1}{n-1}\sum \mathbb{1}[\Delta_i>0]$,
  $\texttt{adj\_dec\_ratio}=\tfrac{1}{n-1}\sum \mathbb{1}[\Delta_i<0]$,
  $\texttt{adj\_eq\_ratio}=\tfrac{1}{n-1}\sum \mathbb{1}[\Delta_i=0]$,
  $\texttt{pos\_delta\_mean/std}$ on $\{\Delta_i>0\}$,
  $\texttt{neg\_delta\_mean/std}$ on $\{\Delta_i<0\}$,
  and $\texttt{sign\_change\_ratio}$ = fraction of sign flips in $(\Delta_1,\ldots,\Delta_{n-1})$.
  \item \textbf{Pairwise order (global monotonicity):}
  $\texttt{pair\_inc\_ratio}=\tfrac{1}{\binom{n}{2}}|\{i<j: v_j>v_i\}|$,
  $\texttt{inversion\_ratio}=\tfrac{1}{\binom{n}{2}}|\{i<j: v_j<v_i\}|$,
  $\texttt{tau\_like}=\texttt{pair\_inc\_ratio}-\texttt{inversion\_ratio}$ (Kendall-tau proxy ignoring ties).
  \item \textbf{Runs and structure:}
  $\texttt{max\_inc\_run}$ = longest strictly increasing contiguous run,
  $\texttt{max\_dec\_run}$ = longest strictly decreasing run,
  $\texttt{num\_monotone\_runs}$ = number of contiguous monotone segments,
  $\texttt{n\_local\_max/min}$ = counts of strict local maxima/minima,
  $\texttt{record\_highs}$ (new maxima count), $\texttt{record\_lows}$ (new minima count).
  \item \textbf{Heuristic LIS approximations (length only):}
  \begin{itemize}
    \item $\texttt{greedy\_len}$: left-to-right “append if $v_i$ increases” record-high count.
    \item $\texttt{greedy\_rev\_len}$: same on the reversed sequence.
    \item $\texttt{beam2}, \texttt{beam3}$: beam-limited LIS lengths with beam $B\in\{2,3\}$.
    \item $\texttt{budget1}, \texttt{budget2}$: greedy with $s\in\{1,2\}$ backtracks (replace tail and truncate at most $s$ times).
  \end{itemize}
  \item \textbf{Patience-sorting descriptors (no direct LIS leakage):}
  From the patience \emph{tails} vector $t$, use
  $\texttt{tail\_mean}$, $\texttt{tail\_std}$, $\texttt{tail\_iqr}$,
  and $\texttt{tail\_slope}$ (OLS slope of $t$ vs.\ index).
  \item \textbf{Reference baseline:}
  $\texttt{rand\_lis\_baseline}=2\sqrt{n}$ (typical LIS scale under random permutations).
\end{itemize}

\paragraph{Pre-processing.}
We replace $\pm\infty$ with NaN and drop rows with missing feature values. Metadata such as $k$ or \texttt{log2\_k} are never used. Standardization is unnecessary for tree ensembles.

\paragraph{Model and hyperparameters.}
We use a Random-Forest Regressor with \texttt{n\_estimators}=800, \texttt{min\_samples\_leaf}=2, \texttt{max\_features}=\texttt{sqrt}, \texttt{random\_state}=42, and \texttt{n\_jobs}=$-1$.

\paragraph{Evaluation and Top-$K$ selection.}
We report $R^2$ and MAE on the held-out group-split test fold. To obtain a compact, interpretable subset, we rank features by RF importance (fit on train) and sweep $K\in\{1,2,3,4,5,6,7,8,9,10,12,15,18,20,25,\ldots\}$. We pick the smallest $K$ whose test performance is within $(\Delta R^2,\Delta\text{MAE})=(0.01,0.02)$ of the full model. We also provide (i) a log-scaled histogram of test residuals and (ii) a Top-$K$ curve (test $R^2$ and MAE vs.\ $K$).


\section{Llama Model Performance}
\label{sec:llama3}

\begin{figure}[t]
    \centering
    \includegraphics[width=\linewidth]{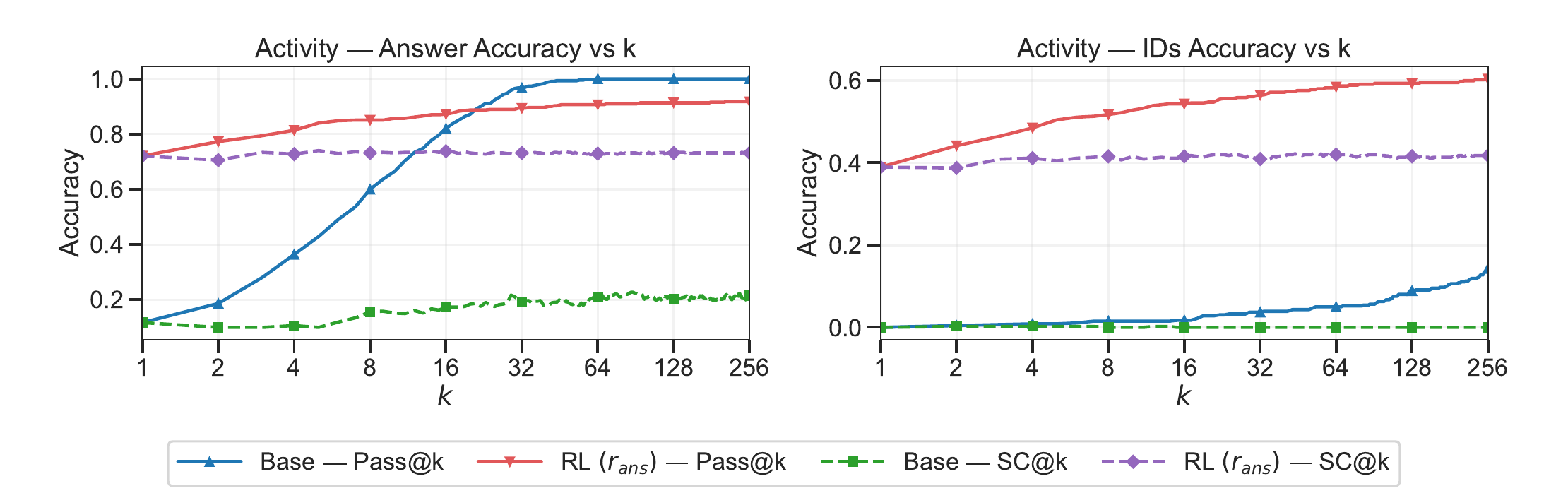}
    \caption{Performance comparison of Base and RLVR($r_{\text{ans}}$) on the \textit{Activity} task with the Llama-3.1-8B model. 
Left: numeric answer accuracy ($Acc_{\text{ans}}$) vs.\ $k$. 
Right: ID sequence accuracy ($Acc_{\text{ids}}$) vs.\ $k$.}
\label{fig:activity_base_vs_rlvr_llama}
\end{figure}

\begin{figure}[t]
    \centering
    \includegraphics[width=\linewidth]{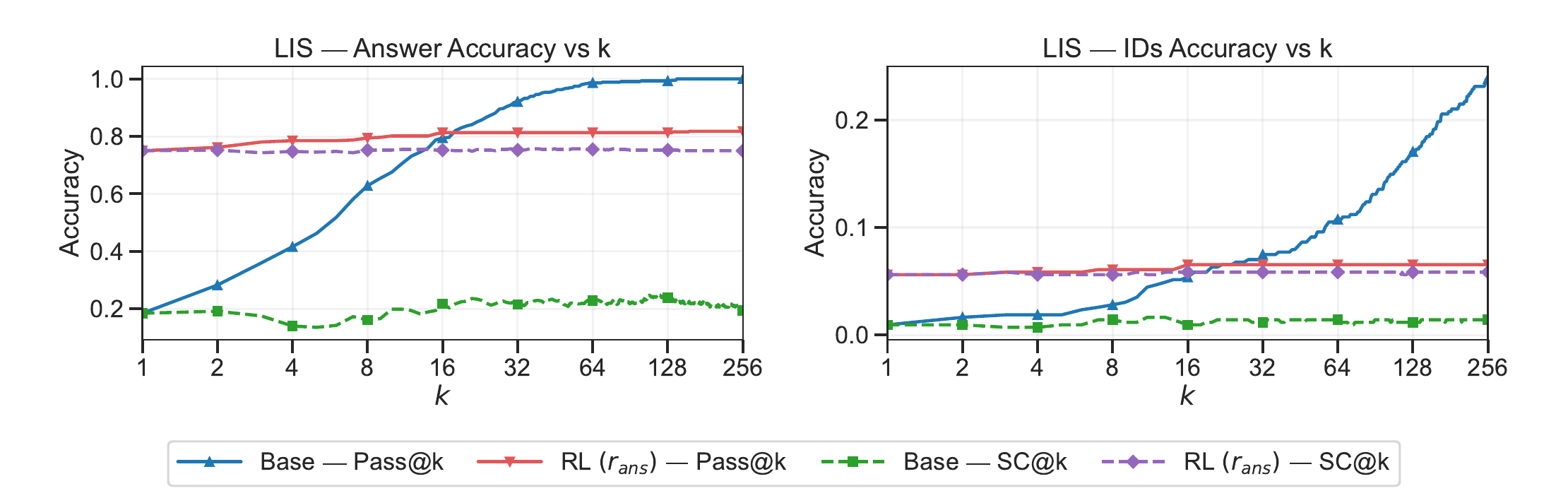}
    \caption{Performance comparison of Base and RLVR($r_{\text{ans}}$) on the \textit{LIS} task with the Llama-3.1-8B model. 
Left: numeric answer accuracy ($Acc_{\text{ans}}$) vs.\ $k$. 
Right: ID sequence accuracy ($Acc_{\text{ids}}$) vs.\ $k$.}
\label{fig:lis_base_vs_rlvr_llama}
\end{figure}

Figure~\ref{fig:activity_base_vs_rlvr_llama} reports results on the \textit{Activity Scheduling} task, and Figure~\ref{fig:lis_base_vs_rlvr_llama} shows results on \textit{LIS} using the \textit{Llama-3.1-8B} model. 
The LLaMA model attains higher overall accuracy on both tasks, as measured by $\mathrm{SC}@256$. 
However, the relative trends between the base and RLVR-trained variants (e.g., under $r_{\text{ans}}$) closely mirror those observed with the Qwen model family.

\begin{figure}[t]
  \centering
  \begin{subfigure}[t]{0.48\linewidth}
    \centering
    \includegraphics[width=\linewidth]{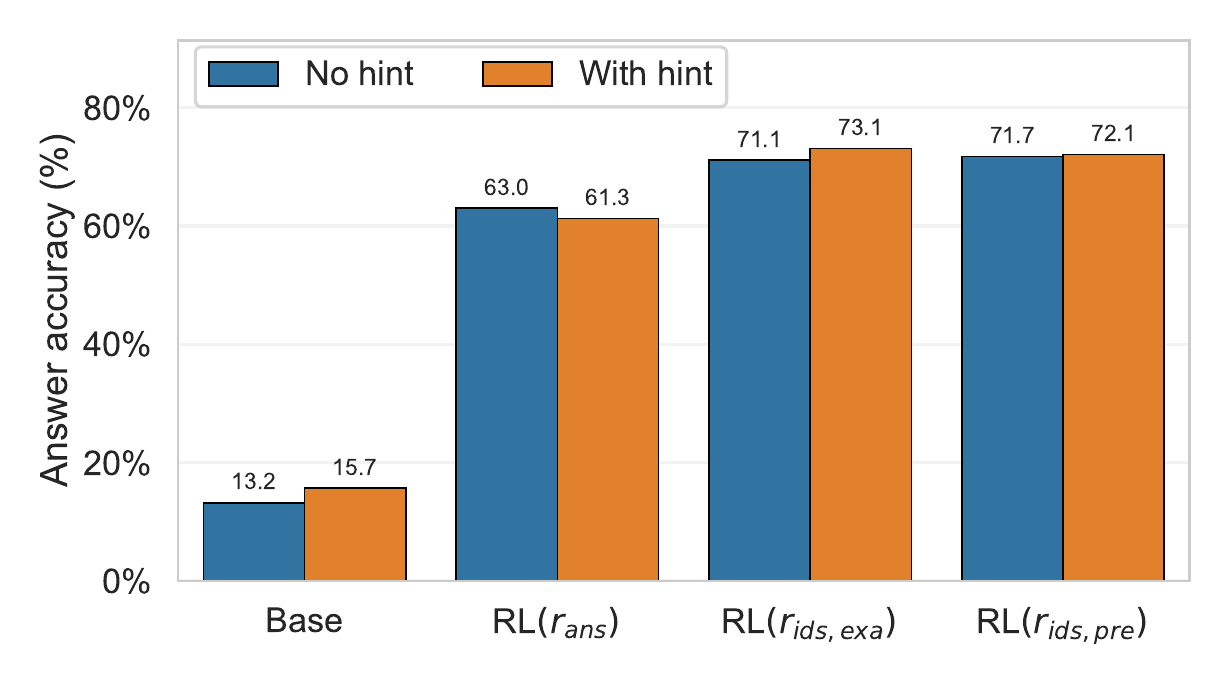}
  \end{subfigure}\hfill
  \begin{subfigure}[t]{0.48\linewidth}
    \centering
    \includegraphics[width=\linewidth]{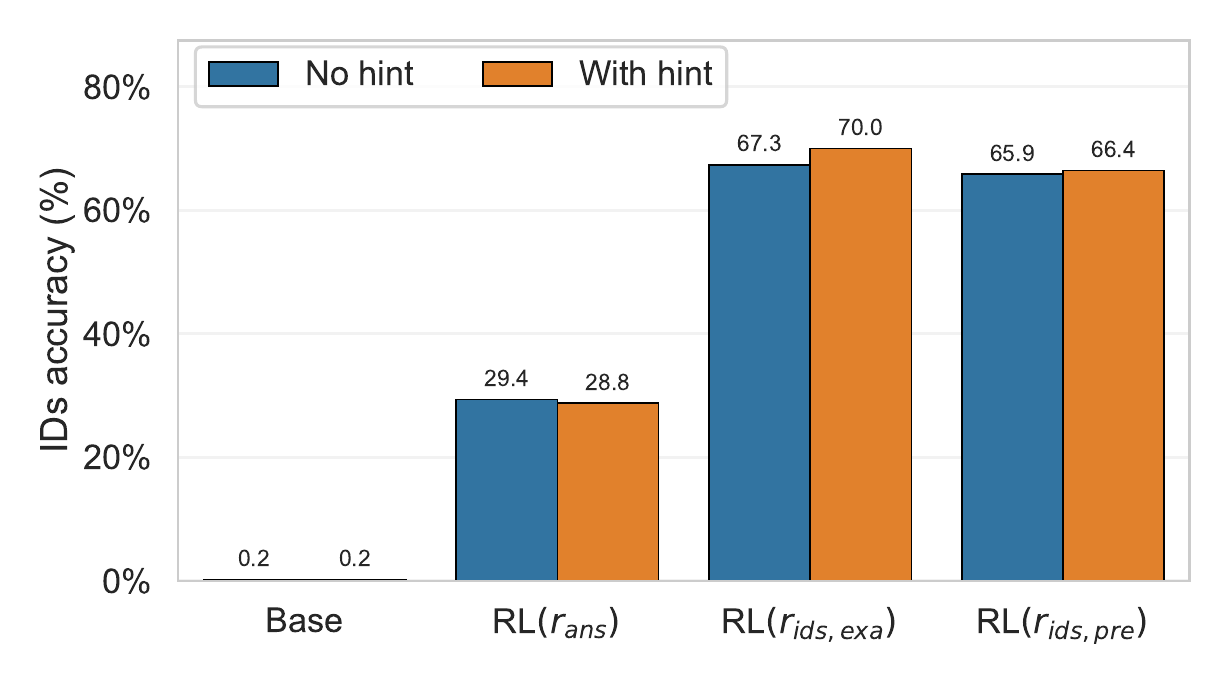}
  \end{subfigure}
  \caption{Acc with/without hint for \textit{Activity.}}
  \label{fig:acc_hint_activity}
\end{figure}

\begin{figure}[t]
  \centering
  \begin{subfigure}[t]{0.48\linewidth}
    \centering
    \includegraphics[width=\linewidth]{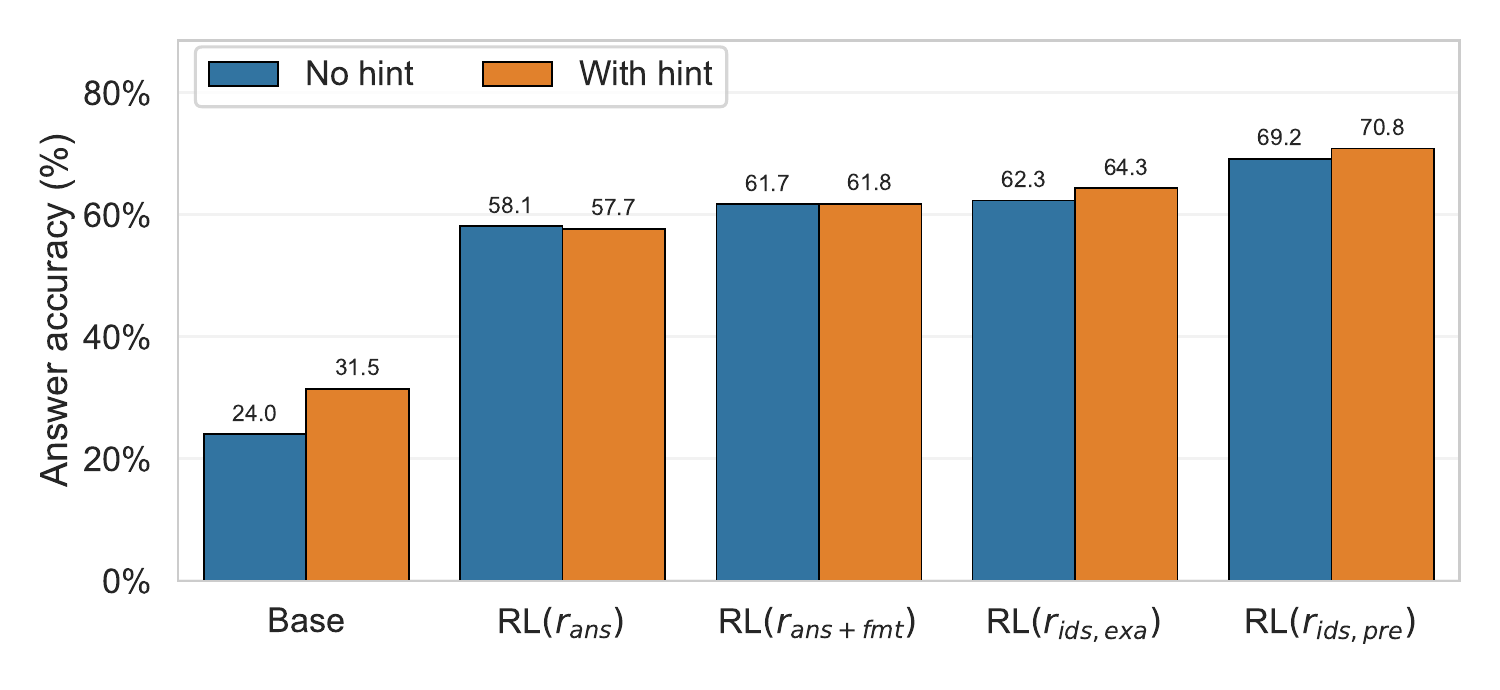}
  \end{subfigure}\hfill
  \begin{subfigure}[t]{0.48\linewidth}
    \centering
    \includegraphics[width=\linewidth]{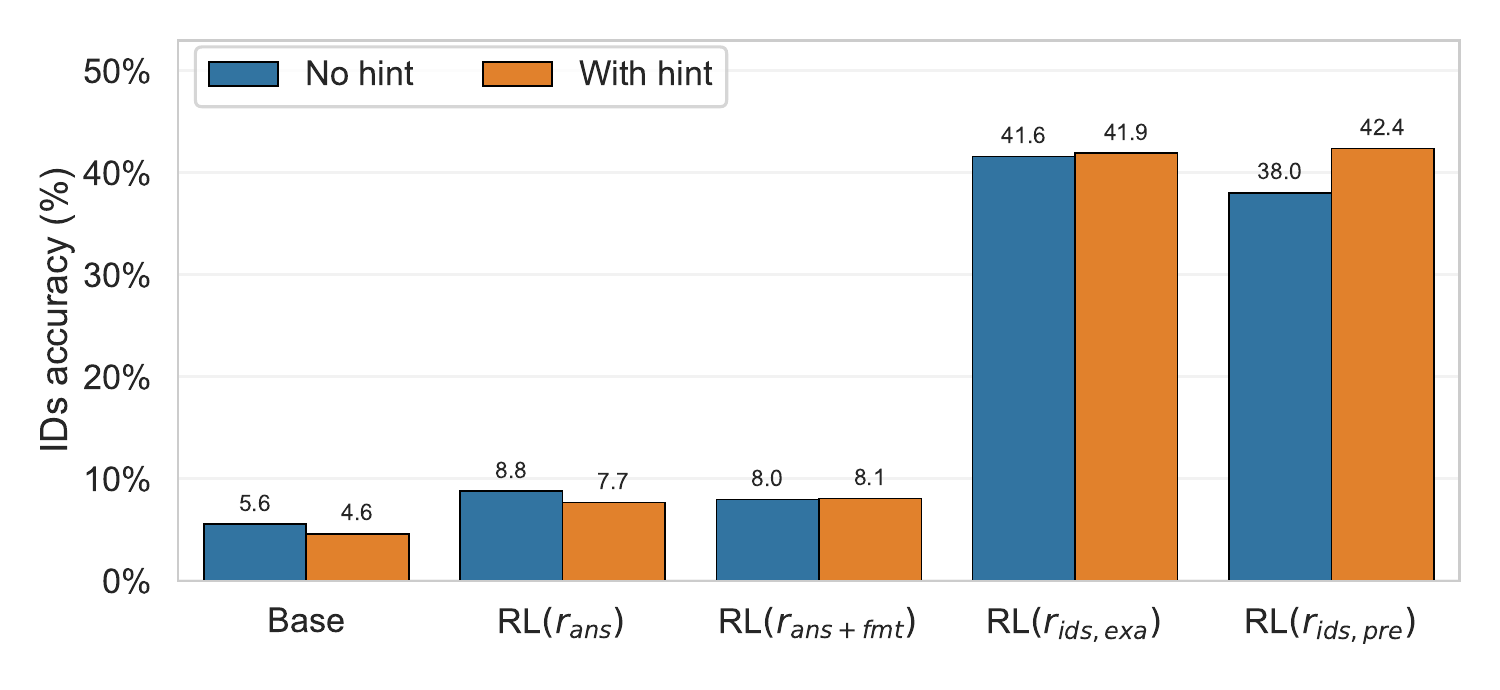}
  \end{subfigure}
  \caption{Acc with/without hint for \textit{LIS.}}
  \label{fig:acc_hint_lis}
\end{figure}

\section{Evaluation of Hinted vs.\ Unhinted Prompts}

Figure~\ref{fig:acc_hint_activity} (Activity) and Figure~\ref{fig:acc_hint_lis} (LIS) compare model performance on test prompts with and without hints. 
Across both tasks, we observe no significant performance differences between the hinted and unhinted variants, suggesting that the models do not substantially benefit from the additional guidance provided by hints.

\end{document}